\documentclass[journal,twoside,web]{ieeecolor}

\usepackage{etoolbox}
\makeatletter
\@ifundefined{color@begingroup}
{\let\color@begingroup\relax
\let\color@endgroup\relax}{}
\def\fix@ieeecolor@hbox#1{%
\hbox{\color@begingroup#1\color@endgroup}}
\patchcmd\@makecaption{\hbox}{\fix@ieeecolor@hbox}{}{\FAILED}
\patchcmd\@makecaption{\hbox}{\fix@ieeecolor@hbox}{}{\FAILED}
\usepackage{tmi}
\usepackage[dvipsnames]{xcolor}
\usepackage{cite}
\usepackage{amsmath,amssymb,amsfonts}
\usepackage{bm}
\usepackage{algorithmic}
\usepackage{graphicx}
\usepackage{textcomp}
\usepackage{orcidlink}
\usepackage{tabularx, booktabs}
\usepackage{multirow, multicol}
\usepackage{makecell}

\hypersetup{
            colorlinks=true,
            linkcolor=blue,
            anchorcolor=blue,
            citecolor=blue
            }

\def\BibTeX{{\rm B\kern-.05em{\sc i\kern-.025em b}\kern-.08em
    T\kern-.1667em\lower.7ex\hbox{E}\kern-.125emX}}
\begin{document}
\title{MedFormer: Hierarchical Medical Vision Transformer with Content-Aware Dual Sparse Selection Attention}
\author{Zunhui Xia$^{\orcidlink{0009-0008-6706-5817}}$, Hongxing Li$^{\orcidlink{0009-0002-7958-3976}}$, and Libin Lan$^{\orcidlink{0000-0003-4754-813X}}$, \IEEEmembership{Member, IEEE}
\thanks{This work described in this paper was supported in part by the Scientific Research Foundation of Chongqing University of Technology under Grant 2021ZDZ030 and in part by the Youth Project of Science and Technology Research Program of Chongqing Education Commission of China under Grant KJQN202301145. \textit{(Corresponding authors: Libin Lan.)}}
\thanks{Zunhui Xia, Hongxing Li, and Libin Lan are with the College of Computer Science and Engineering, Chongqing University of Technology, Chongqing 400054, China (e-mail: zunhui.xia@stu.cqut.edu.cn; hongxing.li@stu.cqut.edu.cn; lanlbn@cqut.edu.cn).}
}

\maketitle

\begin{abstract}
\textit{Objective}. Medical image recognition serves as a key way to aid in clinical diagnosis, enabling more accurate and timely identification of diseases and abnormalities. Vision transformer-based approaches have proven effective in handling various medical recognition tasks. However, these methods encounter two primary challenges. First, they are often task-specific and architecture-tailored, limiting their general applicability. Second, they usually either adopt full attention to model long-range dependencies, resulting in high computational costs, or rely on handcrafted sparse attention, potentially leading to suboptimal performance. To tackle these issues, we present MedFormer, an efficient medical vision transformer with two key ideas. \textit{Approach}. First, it employs a pyramid scaling structure as a versatile backbone for various medical image recognition tasks, including image classification and dense prediction tasks such as semantic segmentation and lesion detection. This structure facilitates hierarchical feature representation while reducing the computation load of feature maps, highly beneficial for boosting performance. Second, it introduces a novel Dual Sparse Selection Attention (DSSA) with content awareness to improve computational efficiency and robustness against noise while maintaining high performance. As the core building technique of MedFormer, DSSA is designed to explicitly attend to the most relevant content. 
Theoretical analysis demonstrates that MedFormer outperforms existing medical vision transformers in terms of generality and efficiency. \textit{Main results}. Extensive experiments across various imaging modality datasets show that MedFormer consistently enhances performance in all three medical image recognition tasks mentioned above.
\textit{Significance}. MedFormer provides an efficient and versatile solution for medical image recognition, with strong potential for clinical application. The code is available on \href{https://github.com/XiaZunhui/MedFormer}{GitHub}.
\end{abstract}

\begin{IEEEkeywords}
Dual sparse selection attention, efficient vision transformer, MedFormer, medical image recognition.
\end{IEEEkeywords}

\section{Introduction}
\label{sec:introduction}
\IEEEPARstart{D}{eep}
learning has been widely applied in biomedical research \cite{le2024hematoma}, \cite{zhao2022improved}, particularly in medical image recognition tasks, such as classification, segmentation, and detection, which are extremely important for computer-aided diagnosis due to their ability to rapidly identify lesion types and accurately localize affected regions \cite{SHAMSHAD2023102802}, \cite{AZAD2024103000}.
Recently, vision transformer-based approaches demonstrate their effectiveness in addressing these medical recognition tasks. 
However, due to the diversity and complexity of medical images across different imaging modalities (e.g., MRI, CT, and Optics) and varying image quality (e.g., signal-to-noise ratio and artifacts), most existing vision transformer architectures are typically tailored for task-specific scenarios, such as disease diagnosis \cite{matsoukas2021time}, \cite{sun2021lesion}, \cite{shao2021transmil}, \cite{zheng2022graph}, anatomical structure and lesion segmentation \cite{chen2021transunet}, \cite{zhang2021transfuse}, \cite{cao2022swin_unet}, \cite{Cai2025BRAU-Net}, \cite{Lan2024BRAU-Net++}, and abnormality detection \cite{tao2021spine}, \cite{wu2023multi}. While effective in their respective applications, such task-specific designs may lead to unsatisfactory performance when applied to diverse medical image recognition tasks. This limitation underscores the need for a general-purpose transformer architecture with task-agnostic, modality-invariant, and quality-adaptive characteristics.
\begin{figure*}
\centering
\includegraphics[width=1\linewidth, keepaspectratio]{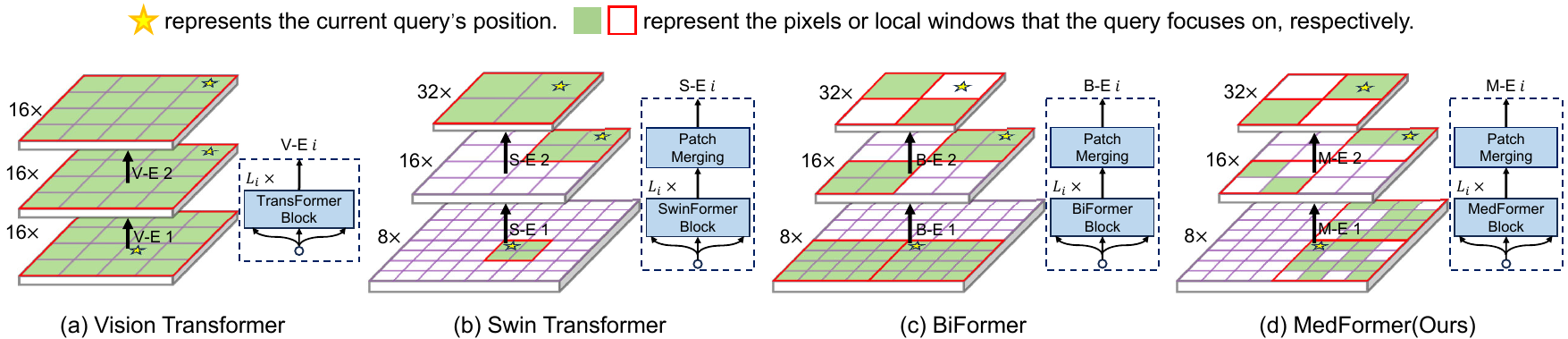}
\caption{
Comparison of different transformer architectures and corresponding attention mechanisms, where ``V-E'', ``S-E'', ``B-E'', and ``M-E'' represent Vision Transformer Encoder, Swin Transformer Encoder, BiFormer Encoder, and MedFormer Encoder, respectively. (a) Vision Transformer \cite{vit} uses full attention, where all layers output feature maps with the same resolution. (b-d) adopt various sparse attention and produce multiscale feature maps. (b) Swin Transformer\cite{liu2021swin} restricts attention computation within local windows. (c) BiFormer\cite{zhu2023biformer} dynamically selects windows for each query. (d) Our proposed MedFormer adaptively allocates attention at both region-level and pixel-level to reduce computational complexity and extract more effective features. This hierarchical architecture improves performance in various dense prediction tasks.}
\label{fig1:motivation}
\end{figure*} 

On the other hand, as the core building block of vision transformers, attention is used to model long-range dependencies. When applied directly to medical imaging tasks, vanilla attention incurs high computational complexity and a heavy memory footprint \cite{chen2021transunet}, \cite{zhang2021transfuse}, \cite{zheng2021rethinking}. 
To alleviate this issue, recent contributions \cite{cao2022swin_unet}, \cite{Lan2024BRAU-Net++}, \cite{medt} focus on adapting the sparse attention  to construct various transformer architectures that efficiently handle medical vision tasks.
However, some of them such as Swin-Unet \cite{cao2022swin_unet} and MedT \cite{medt} typically adopt static, handcrafted sparse attention to select patterns.
Another line of works like BRAU-Net \cite{Cai2025BRAU-Net} and its enhanced version BRAU-Net++ \cite{Lan2024BRAU-Net++}, drawn inspiration from  Bi-level Routing Attention (BRA) \cite{zhu2023biformer}, use dynamic and content-aware mechanism to select the most relevant content. 
The former enables each query token to attend to its designated neighboring tokens, limiting the ability to capture long-range dependencies. The latter allows each query token to attend to a small portion of the most semantically relevant tokens selected within a larger scale. While more beneficial for achieving the speed-accuracy trade-off, this line of work essentially makes a sparse selection of region-level tokens derived from the average of all pixels within each region, which may inadvertently incorporate pixel-level noise tokens into the computation of the pixel-level attention matrix, potentially resulting in performance degradation in scenarios with high intra-region variability. Therefore, dynamically adjusting the selection criteria for the most relevant tokens from a pixel-level perspective is essential to improve model robustness and accuracy by mitigating noise incorporation and capturing pertinent tokens. 

In this work, inspired by the successful pyramid vision transformers\cite{liu2021swin}, \cite{zhu2023biformer}, \cite{wang2021PVT},  we propose a unified medical vision transformer backbone, called \textbf{MedFormer}, which is constructed in a pyramid structure suitable for various medical applications, such as disease diagnosis, anatomical structure and lesion segmentation, and abnormality detection. Various vision transformer backbone, including the proposed MedFormer, are shown in Fig. \ref{fig1:motivation}.
This structure has the following several advantages. First, this hierarchical design can reduce the computational cost associated with processing feature maps at different resolutions while yielding richer semantic information beneficial for dense prediction. Second, MedFormer facilitates the implementation of our adapted attention mechanism on feature maps across different scales. This flexibility allows the model to apply attention selectively based on the resolution of the feature map. In high-resolution feature maps, attention can focus more on local regions, which is crucial for tasks like lesion segmentation, while in low-resolution maps, it emphasizes global information, aiding in tasks such as disease diagnosis. Third, MedFormer increases channel depth while reducing spatial dimensions. This strategy maintains rich feature representations even at lower resolutions, ensuring that the model retains sufficient detail for precise segmentation and detection tasks.

In addition to advances in model architecture, we introduce a novel, dynamic, and content-aware sparse attention aimed at achieving both higher model performance and computational efficiency. Unlike previous approaches such as full attention \cite{vit}, local attention \cite{liu2021swin}, and bi-level routing attention \cite{zhu2023biformer}, our proposed sparse attention selectively attends to the most relevant content at both region-level \cite{zhu2023biformer} and pixel-level \cite{zhao2019explicit} scales, further reducing computational cost and excluding the noise pixels without sacrificing performance. We refer to this attention mechanism as \textbf{D}ual \textbf{S}parse \textbf{S}election \textbf{A}ttention (\textbf{DSSA}), since it explicitly performs twice sparse selection -- once at the region level and once at the pixel level. Specifically, it first filters out irrelevant key-value regions through region-level sparse selection, retaining only the $k_1$ most relevant regions. Subsequently, it performs pixel-level sparse selection within these selected regions to further exclude irrelevant pixels (including noise pixels) and retain the most pertinent $k_2$ pixels. This dual sparse selection approach ensures that the model attends to the most significant features while significantly reducing computational complexity. After the dual sparse selection operations, pixel-to-pixel  full attention is applied to the regathered pixel-level tokens from the second pixel-level sparse selection step. 
Various attention mechanisms, including the proposed DSSA, are illustrated in Fig. \ref{fig1:motivation}. The designed DSSA has the following several advantages. First, DSSA reduces the number of tokens involved in the attention computation by performing two-step sparse selection, thereby lowering computational complexity. Second, DSSA ensures the model focuses on critical information by attending to the most relevant region-level and pixel-level tokens, avoiding the negative impact of pixel-level noise tokens and leading to improved performance and efficiency. Third, DSSA can be effectively implemented on feature maps across different resolutions, making it highly suitable for our MedFormer network structure.

The main technical contributions---hierarchical feature representations through a pyramid structure and the DSSA mechanism---endow MedFormer with a strong ability to effectively model long-range dependencies. As a result, the proposed MedFormer architecture serves as a robust general-purpose backbone without any additional bells and whistles. This makes it particularly appropriate for medical recognition tasks such as disease diagnosis, anatomical structure segmentation, and abnormality detection.
To the best of our knowledge, this is the first work to attempt developing a general-purpose vision transformer backbone for the medical image recognition community.
Overall, our main contributions can be summarized as follows:

\begin{enumerate}
    \item We propose an effective and efficient medical vision transformer, named MedFormer, which is constructed hierarchically to generate multiscale feature representations. MedFormer is capable of serving as a general-purpose backbone for various medical visual recognition tasks, including classification and dense prediction tasks such as segmentation and detection.
    \item We design a dual sparse selection attention mechanism, termed DSSA, which performs one-time sparse selection on both region-level and pixel-level scales, respectively. This mechanism offers two significant advantages: it capably selects the most relevant tokens while reducing computational cost, and it effectively improves robustness against noise tokens.
    \item We provide a detailed theoretical analysis of DDSA and evaluate the proposed MedFormer across a variety of medical visual recognition tasks, demonstrating superior performance compared to other state-of-the-art methods under almost all evaluation metrics.
\end{enumerate}

The rest of this paper is organized as follows. 
Section \hyperref[sec:relatedwork]{\uppercase\expandafter{\romannumeral2}} delves into the prior related works and their limitations.
Section \hyperref[sec:method]{\uppercase\expandafter{\romannumeral3}} specifies our method, including MedFormer architecture and DSSA mechanism. Section \hyperref[sec:exp-results]{\uppercase\expandafter{\romannumeral4}} showcases the experimental results and details, particularly the visualization of DSSA attention map. Section \hyperref[sec:conclusion]{\uppercase\expandafter{\romannumeral5}} summarizes our findings and indicates directions for further research.

\begin{figure*}
\centering
\includegraphics[width=1\linewidth, keepaspectratio]{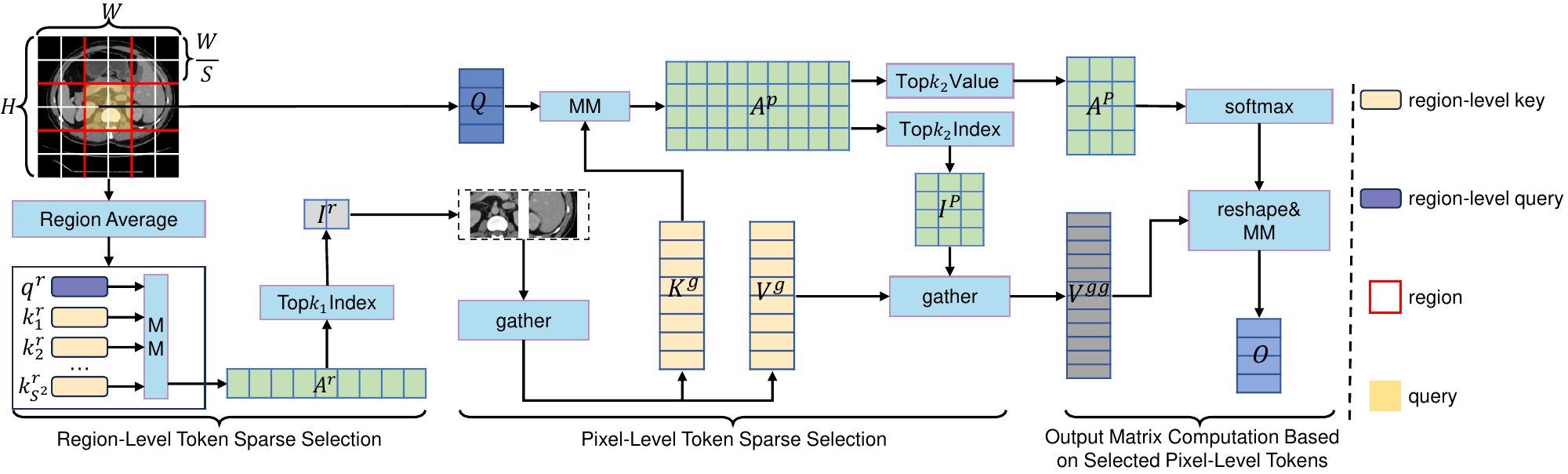}
\caption{Illustration of dual sparse selection attention (DSSA). First, region-level token sparse selection is performed by obtaining queries and key features through region averaging, selecting the top-$k_1$ relevant regions. Then, pixel-level token sparse selection is conducted within these selected regions, choosing the top-$k_2$ relevant pixel tokens. Finally, matrix multiplication is applied to obtain the output. The mechanism can reduce computational complexity while extracting the most relevant features.}
\label{fig2:block}
\end{figure*}

\section{Related Work}
\label{sec:relatedwork}
\subsection{Transformer-based Medical Visual Recognition}
\label{sec:medical visual recognition}
Due to the outstanding performance in modeling global context, transformers have achieved unprecedented success in the natural language processing and computer vision tasks. Given this phenomenal success, transformers have also been adapted to medical visual recognition field, achieving competitive or even superior results compared to CNN counterparts across diverse modalities like MRI, CT, and X-ray and various tasks such as classification, segmentation, and detection.

During the early phase of adaptation, several main works have emerged that apply vanilla transformers to medical visual recognition tasks. For classification, notable approaches include MIL-VT \cite{yu2021mil-vt} and SEViT \cite{Faris2022SEViT}. Segmentation has benefited from models like TransUNet \cite{chen2021transunet}, TransFuse \cite{zhang2021transfuse}, and BAT \cite{wang2021boundary}. Detection tasks have seen advancements with Spine-Transformers \cite{tao2021spine} and TR-Net \cite{MA2021TR-Net}. 
However, these methods typically employ full attention to capture long-range dependencies, which leads to high computational complexity that scales quadratically with sequence length. 
Inspired by pioneering works such as PVT \cite{wang2021PVT}, Swin Transformer \cite{liu2021swin}, and BiFormer \cite{zhu2023biformer}, we propose an efficient medical vision transformer that leverages a well-designed dual sparse selection attention mechanism. This novel architecture serves as a general-purpose backbone for various medical image recognition tasks.

\subsection{Efficient Self-Attention in Medical Visual Recognition}
As noted above, early adaptations to the medical visual recognition field typically adopted full attention to capture long-range dependencies. However, a well-known concern with this approach is its quadratic computational complexity, which can hinder model scalability in many settings. To mitigate this issue, more recent works have concentrated on developing modifications to full attention aimed at improving computational efficiency while maintaining or even enhancing performance. Regarding the classification task, IDA-Net \cite{ZHAO2023IDA-Net} incorporates deformable attention \cite{Xia2022deformable_attention}, allowing for flexible and adaptive sampling of features. Inspired by Swin Transformer \cite{liu2021swin}, HiFuse \cite{HUO2024105534} designs a three-branch hierarchical multi-scale feature fusion network structure with rich scalability and linear computational complexity, achieving good classification performance. POCFormer \cite{perera2021pocformer}, which exploits low-rank approximations \cite{wang2020linformer} to reduce the complexity of vanilla attention from quadratic to linear for COVID-19 detection.   
KAT \cite{Zheng2023KAT} and MG-Trans \cite{Shi2023MG-Trans} respectively utilize kernelization techniques and graph-enhanced transformers, effectively compressing the attention mechanism to handle longer sequences more efficiently for histopathology image classification. 
For segmentation tasks, MedT \cite{medt} and MSAANet \cite{zeng2023msaanet} restrict query attention key-value pairs in both width and height dimensions to reduce computational burden. Similarly, Swin-Unet \cite{cao2022swin_unet}, HiFormer \cite{heidari2023hiformer}, MAXFormer \cite{liang2023maxformer} perform attention computation within local windows. 
BRAU-Net++ \cite{Lan2024BRAU-Net++} leverages dynamic sparse attention mechanism \cite{zhu2023biformer}, focusing on selective regions of interest to streamline computations without sacrificing accuracy. Furthermore, numerous studies have explored the use of various efficient attention mechanisms specifically tailored for segmentation tasks. One such example is MISSFormer \cite{huang2022missformer}, which introduces an efficient self-attention module to model long-range dependencies while capturing local context from multi-scale features. Different from these methods, inspired by bi-level routing attention \cite{zhu2023biformer} and explicit sparse attention \cite{zhao2019explicit}, we propose dual sparse selection attention for various medical image recognition tasks, which not only reduces computational and memory costs but also improves performance and robustness to noise tokens.

\subsection{Transformer-based Medical Vision Backbone}
Recently, transformer-based backbones become a dominant paradigm in the computer vision community. The pioneering work is Vision Transformer (ViT) \cite{vit}, which achieves remarkable success in image classification. However, ViT is a columnar structure, leading to low-resolution outputs. To address this limitation, Pyramid Vision Transformer (PVT) \cite{wang2021PVT} borrows the pyramid structure from CNN and proposes a similar structure as the general vision transformer backbone. Inspired by the significant achievements of PVT in a wide range of vision tasks, several subsequent studies such as Swin Transformer \cite{liu2021swin},  BiFormer \cite{zhu2023biformer}, DAT \cite{Xia2022deformable_attention} have also embraced this architecture and enhanced it with their respective efficient attention mechanisms specifically designed for dense prediction tasks. Similarly, our work also proposes a four-stage pyramid structure that is equipped with well-designed dual sparse selection attention to address dense prediction tasks in the medical image recognition field, 
which enjoys good model performance and high computational efficiency.

\begin{figure*}
\centering
\includegraphics[width=1\linewidth]{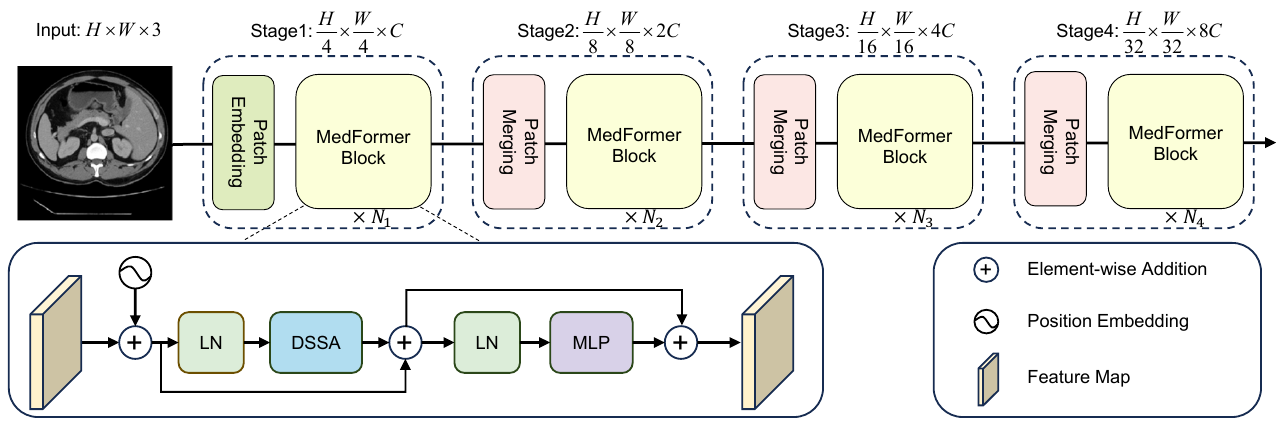}
\caption{\textbf{Top}: The overall architecture of MedFormer is illustrated, presenting a four-stage hierarchical pyramid network. At each stage $i$, the input feature maps undergo patch embedding or patch merging layer to reduce spatial resolution while increasing the number of channels. Each stage contains $N_i$ stacked MedFormer Blocks. This design helps to extract multiscale features while maintaining computational efficiency. \textbf{Bottom}: Detailed design of the MedFormer Block.}
\label{fig3:backbone}
\end{figure*}

\section{Method}
\label{sec:method}
In this section, we first introduce the design details of the DSSA mechanism. Then, we theoretically analyze its computational complexity. Finally, we build the DSSA-based MedFormer Block and propose a novel backbone network for medical image recognition, called MedFormer.
\subsection{Dual Sparse Selection Attention (DSSA)}
The self-attention mechanism is capable of capturing global context, but its computational complexity increases quadratically with the length of input sequence. Consequently, when processing longer sequences, the self-attention mechanism may become a bottleneck for Transformer. To overcome this limitation, we propose the content-aware DSSA. A conceptual illustration of the DSSA is provided in Fig. \ref{fig2:block}. Specifically, our approach adopts dynamic sparse selection at both the regional and pixel level, which substantially reduces computational complexity while filtering noise tokens to ensure effective feature extraction. To achieve this, the input feature map $\mathbf{X} \in \mathbb{R}^{H \times W \times C}$ is partitioned into $S \times S$ non-overlapping regions, represented as $\mathbf{X^r} \in \mathbb{R}^{S^2 \times \frac{HW}{S^2} \times C}$. This transformation preserves spatial relationships among pixels, enabling the direct computation of partitioned queries $\mathbf{Q}\in \mathbb{R}^{S^{2}\times\frac{HW}{S^{2}}\times C}$, keys $\mathbf{K}\in \mathbb{R}^{S^{2}\times\frac{HW}{S^{2}}\times C}$,  and values $\mathbf{V}\in \mathbb{R}^{S^{2}\times\frac{HW}{S^{2}}\times C}$ by the projection weight matrices $\mathbf{W}^q$, $\mathbf{W}^k$, $\mathbf{W}^v \in \mathbb{R}^{C \times C}$, this process is formularized as: 
\begin{equation}\mathbf{Q}=\mathbf{X^r}\mathbf{W}^q,\mathbf{K}=\mathbf{X^r}\mathbf{W}^k,\mathbf{V}=\mathbf{X^r}\mathbf{W}^v.\end{equation}

\subsubsection{Region-Level Token Sparse Selection}
Directly performing pixel-level sparse selection is not only time-consuming but also computationally expensive. To address this challenge, we adopt a region-level sparse selection strategy, as the number of regions is substantially smaller than the number of pixels, thereby significantly reducing computational complexity. Specifically, we calculate the average of queries and keys within each region, obtaining region-level queries $\mathbf{Q}^r\in \mathbb{R}^{S^{2} \times C}$ and region-level keys $\mathbf{K}^r \in \mathbb{R}^{S^{2} \times C}$. Then, we compute the relevance between all region-level tokens based on $\mathbf{Q}^r$ and $\mathbf{K}^r$.
\begin{equation}\mathbf{A}^r=\mathbf{Q}^r(\mathbf{K}^r)^T.\end{equation}
Where, $\mathbf{A}^r \in \mathbb{R}^{S^{2} \times S^{2}}$ reflects the semantic relevance between all region-level tokens. To efficiently and adaptively filter region-level tokens, we retain the $k_1$ regions with the highest relevance to the query region. The indices of these $k_1$ region-level tokens are stored in the set $\mathbf{I}^{r}\in \mathbb{N}^{S^{2}\times k_{1}}$ and are shared across all pixel-level tokens within the corresponding query region. This approach dynamically reduces the number of regions while preserving the capability for global modeling.
\begin{equation}\mathbf{I}^r=\mathrm{Topk_1Index}(\mathbf{A}^r).\end{equation}

\subsubsection{Pixel-Level Token Sparse Selection}
To exploit the GPU's parallel computation capabilities for matrices, we gather the pixel-level keys and values distributed across $k_1$ regions into new tensors, denoted as $\mathbf{K}^{g}\in \mathbb{R}^{S^{2}\times\frac{k_{1}HW}{S^{2}}\times C}$ and $\mathbf{V}^{g}\in \mathbb{R}^{S^{2}\times\frac{k_{1}HW}{S^{2}}\times C}$.
\begin{equation}
    \mathbf{K}^{g}=\mathrm{gather}(\mathbf{K},\mathbf{I}^r),\mathbf{V}^{g}=\mathrm{gather}(\mathbf{V},\mathbf{I}^r).
\end{equation}

Notably, not all selected pixel-level tokens contribute beneficial features to model learning, as the region-level sparse selection process operates at a coarse granularity. To further eliminate  noise tokens and reduce computational overhead, we propose a pixel-level sparse selection strategy. Similar to the region-level selection process, we calculate the relevance between the selected pixel-level tokens and the corresponding query, retaining the $k_2$ pixel-level tokens with the highest relevance to the query. The values and indices of these pixel-level tokens are stored in $\mathbf{A}^{p}\in \mathbb{R}^{S^{2}\times\frac{HW}{S^{2}}\times\frac{k_{1}HW}{S^{2}}}$ and $\mathbf{I}^{p}\in \mathbb{N}^{S^{2}\times{\frac{HW}{S^{2}}}\times k_{2}}$, respectively, where the value of 
 $k_2$ relates to a scaling factor 
$\lambda$. The specific formulas are as follows:
\begin{equation}
    k_2=\lambda\frac{k_{1}HW}{S^{2}},
\end{equation}
\begin{equation}
\mathbf{A}^p=\mathbf{Q}(\mathbf{K}^g)^T,
\end{equation}
\begin{equation}
    \mathbf{A}^{P}=\mathrm{Topk_2Value}(\mathbf{A}^{p}),
\end{equation}
\begin{equation}
    \mathbf{I}^{P}=\mathrm{Topk_2Index}(\mathbf{A}^{p}).
\end{equation}

\subsubsection{Matrix Multiplication}
By leveraging the dual sparse selection mechanism, we efficiently complete the attention allocation process. Then, we gather the $k_2$ pixel-level values into a new tensor $\mathbf{V}^{gg}\in \mathbb{R}^{S^{2}\times\frac{HW}{S^{2}}\times k_{2}\times C}$, to fully exploit GPU acceleration:
\begin{equation}
    \mathbf{V}^{gg}=\mathrm{gather}(\mathbf{V}^g,\mathbf{I}^P).
\end{equation}

To enable matrix multiplication between $\mathbf{A}^{P}$ and $\mathbf{V}^{gg}$, we introduce an additional dimension to the third axis of $\mathbf{A}^{P}$, resulting in a tensor shape of $S^{2} \times \frac{HW}{S^{2}} \times 1 \times k_2$. Furthermore, to enrich local feature information, we incorporate a local context enhancement module, denoted as $\mathrm{LCE}(\cdot)$, which is constructed using a 5$\times$5 depth-wise convolution. This module further strengthens feature representation. The final output of the model is expressed as:
\begin{equation}\mathbf{O}=\mathbf{A}^{P}\mathbf{V}^{gg}+\mathrm{LCE}(\mathbf{V}).\end{equation}

\begin{table*}
\centering
\small
\caption{Variants of MedFormer for medical image classification. $C_i$: Channel dimension at stage $i$. $S$: the number of regions in feature maps. $k_1$: Select the $k_1$ most relevant regions. $\lambda$: proportionality coefficient used for the second sparse selection.}
\label{tab1:version}
\resizebox{1.0\linewidth}{!}{
    \begin{tabular}{p{1.6cm}<{\centering} | p{1.6cm}<{\centering} |p{2.9cm}<{\centering} |p{3cm}<{\centering}| p{3cm}<{\centering}| p{3cm}<{\centering}}
         \toprule
    Stage & Output Size & Layer & MedFormer-Tiny & MedFormer-Small & MedFormer-Base \\
    \midrule
    \multirow{2}{*}[-3ex]{Stage1} & \multirow{2}{*}[-3ex]{$\frac{H}{4} \times \frac{W}{4}$}  & Patch Embedding & 
    $\begin{bmatrix} 3 \times 3 \\ \mathrm{BN} \end{bmatrix} \times 2$, $C_1=32$ & 
    $\begin{bmatrix} 3 \times 3 \\ \mathrm{BN} \end{bmatrix} \times 2$, $C_1=64$ & 
    $\begin{bmatrix} 3 \times 3 \\ \mathrm{BN} \end{bmatrix} \times 2$, $C_1=96$ \\
    \cmidrule{3-6}
     & & Transformer Encoder & \multicolumn{3}{c}{$\begin{bmatrix} S=7 \\ k_1=1 \\ \lambda=\frac{1}{8} \end{bmatrix} \times 2$} \\
    \midrule
    \multirow{2}{*}[-3ex]{Stage2} & \multirow{2}{*}[-3ex]{$\frac{H}{8} \times \frac{W}{8}$}  & Patch Merging& 
    $\begin{bmatrix} 3 \times 3 \\ \mathrm{BN} \end{bmatrix}$, $C_2=64$  & 
    $\begin{bmatrix} 3 \times 3 \\ \mathrm{BN} \end{bmatrix}$, $C_2=128$ & 
    $\begin{bmatrix} 3 \times 3 \\ \mathrm{BN} \end{bmatrix}$, $C_2=192$ \\
    \cmidrule{3-6}
    & & Transformer Encoder & \multicolumn{3}{c}{$\begin{bmatrix} S=7 \\ k_1=4 \\ \lambda=\frac{1}{8} \end{bmatrix} \times 2$} \\
    \midrule
    \multirow{2}{*}[-3ex]{Stage3} & \multirow{2}{*}[-3ex]{$\frac{H}{16} \times \frac{W}{16}$}  & Patch Merging & 
    $\begin{bmatrix} 3 \times 3 \\ \mathrm{BN} \end{bmatrix}$, $C_2=128$  & 
    $\begin{bmatrix} 3 \times 3 \\ \mathrm{BN} \end{bmatrix}$, $C_2=256$ & 
    $\begin{bmatrix} 3 \times 3 \\ \mathrm{BN} \end{bmatrix}$, $C_2=384$ \\
    \cmidrule{3-6}
    & & Transformer Encoder & \multicolumn{3}{c}{$\begin{bmatrix} S=7 \\ k_1=16 \\ \lambda=\frac{1}{8} \end{bmatrix} \times 8$} \\
    \midrule
    \multirow{2}{*}[-3ex]{Stage4} & \multirow{2}{*}[-3ex]{$\frac{H}{32} \times \frac{W}{32}$}  & Patch Merging & 
    $\begin{bmatrix} 3 \times 3 \\ \mathrm{BN} \end{bmatrix}$, $C_2=256$  & 
    $\begin{bmatrix} 3 \times 3 \\ \mathrm{BN} \end{bmatrix}$, $C_2=512$ & 
    $\begin{bmatrix} 3 \times 3 \\ \mathrm{BN} \end{bmatrix}$, $C_2=768$ \\
    \cmidrule{3-6}
    & & Transformer Encoder & \multicolumn{3}{c}{$\begin{bmatrix} S=7 \\ k_1=49 \\ \lambda=\frac{1}{8} \end{bmatrix} \times 2$} \\
    \midrule
    \multicolumn{3}{c|}{Params} & 3.20M & 12.63M & 28.30M \\
    \multicolumn{3}{c|}{FLOPs} & 0.58G & 2.18G & 4.79G \\
    \bottomrule
    \end{tabular}
}
\end{table*}

\subsection{Complexity Analysis of DSSA}
In this section, we analyze the computational complexity of DSSA in terms of FLOPs. The total FLOPs are composed of three primary components: linear projection, region-level token sparse selection, and pixel-level token sparse selection. Therefore, the total FLOPs can be represented as follows:

\begin{equation*}
\mathrm{FLOPs}_{proj}= \underbrace{3HWC^2}_{\mathrm{projection}}
\end{equation*}
\begin{equation*}
\mathrm{FLOPs}_{region}=\underbrace{ S^2(S^2C)}_{\mathbf{Q}^r \cdot (\mathbf{K}^r)^T}+\underbrace{S^2(S^2+k_1)}_{\mathrm{Topk_1}}\\
\end{equation*}
\begin{equation*}
\mathrm{FLOPs}_{token}=\underbrace{HW\frac{{k_1}HW}{S^2}C}_{\mathbf{Q}\cdot (\mathbf{K}^g)^T}+\underbrace{HW(\frac{{k_1}HW}{S^2}+k_2)}_{\mathrm{Topk_2}}+\underbrace{HW{k_2}C}_{\mathbf{A}^{P}\cdot \mathbf{V}^{gg}}
\end{equation*}
\begin{equation}\label{eq:complexity}
\begin{aligned}
    \mathrm{FLOPs} &= \mathrm{FLOPs}_{proj} + \mathrm{FLOPs}_{region} + \mathrm{FLOPs}_{token}\\
         &=3HWC^2 + S^2(S^2C+S^2+k_1)\\
         &+ HW(C+1)(\frac{{k_1}HW}{S^2}+k_2),
\end{aligned}
\end{equation}
where $S^2C > S^2 + k_1 $ and $ \frac{{k_1}HW}{S^2} > k_2 $, we then substitute $S^2C$ and $ \frac{{k_1}HW}{S^2}$ into Eq. \ref{eq:complexity} to obtain the inequality:
\begin{equation}
\label{eq:inequality1}
    \mathrm{FLOPs}< 3 HWC^2+2(S^2)^{2}C+\frac{2k_1(HW)^2}{S^2}(C+1).
\end{equation}
Next, we denote the right hand side of the inequality in Eq. \ref{eq:inequality1} as $\mathrm{InEq}$. Applying the AM-GM inequality $a+b+c \geq 3(abc)^{\frac{1}{3}}$, we proceed with further analysis:
\begin{equation}\label{eq:inequality2}
\begin{aligned}
    \mathrm{InEq}& = 3 HWC^2 + 2(S^2)^{2}C + \frac{2k_1(HW)^2}{S^2}(C + 1) \\
    & > 3 HWC^2 + 2(S^2)^{2}C + \frac{2k_1(HW)^2}{S^2}C \\
    & = 3 HWC^2 + 2(S^2)^{2}C + \frac{k_1(HW)^2}{S^2}C + \frac{k_1(HW)^2}{S^2}C \\
    &\geq 3HWC^2 + 3\left(2(S^2)^2C \cdot \frac{k_1(HW)^2}{S^2}C \cdot \frac{k_1(HW)^2}{S^2}C\right)^{\frac{1}{3}} \\
    &= 3HWC^2 + 6C(k_1)^{\frac{2}{3}}(HW)^{\frac{4}{3}}.
\end{aligned}
\end{equation}
Combining Eq. \ref{eq:inequality1} and Eq. \ref{eq:inequality2}, it is evident that the upper bound of FLOPs is strictly less than the lower bound of $\mathrm{InEq}$:
\begin{equation}
    \mathrm{FLOPs}<3HWC^2+6C(k_1)^{\frac{2}{3}}(HW)^{\frac{4}{3}}.
\end{equation}

In conclusion, this analysis demonstrates that the complexity of DSSA is less than $O((HW)^{\frac{4}{3}})$, representing a significant reduction compared to vanilla attention.

\begin{table*}[htpb]
\centering
\caption{Datasets of the various medical image tasks used in our experiments.}
\resizebox{1.0\linewidth}{!}
{

\begin{tabular}{p{2.1cm} | p{4.4cm}<{\centering} |p{2cm}<{\centering}| p{1.3cm}<{\centering}| p{1.3cm}<{\centering} p{1.3cm}<{\centering} p{1.3cm}<{\centering} }
        \toprule
        Task & DataSet & Input\_size & Num\_class & Train & Val & Test \\
        \midrule
        \multirow{3}{*}[-0.3ex]{Classification} & ISIC-2018 Classification\cite{tschandl2018ham10000}\cite{codella2019skin} & 224$\times$224 & 7 & 9376 & - & 2344 \\
        & ColonPath \cite{MedFMC} & 224$\times$224 & 2 & 8007 & - & 2002 \\
        & Brain Tumor \cite{brain_tumor_classification} & 224$\times$224 & 4 & 5618 & - & 1405 \\
        \midrule
        \multirow{3}{*}[-0.3ex]{Segmentation} & ISIC-2018 Segmentation \cite{tschandl2018ham10000}\cite{codella2019skin} & 224$\times$224 & 2 & 1868 & 467 & 259 \\
        & CVC-ClinicDB \cite{cvc} & 256$\times$256 & 2 & 441 & 110 & 61 \\
        & Synapse \cite{Synapse} & 224$\times$224 & 9 & 2212 & - & 1567 \\
        \midrule
        \multirow{2}{*}[0.1ex]{Detection} & Kvasir-Seg \cite{kvasir} & 256$\times$256 & 2 & 800 & - & 200 \\
        & Brain Tumor Detection \cite{brain_tumor_detection} & 256$\times$256 & 4 & 7128 & - & 1782\\
        \bottomrule
    \end{tabular}
}
\label{tab2:data}
\end{table*}

\begin{table*}[h]
    \centering 
   \caption{Comparison of different methods on three medical image classification datasets. * indicates that the model is initialized with weights pretrained on the ImageNet dataset.}
    \resizebox{1.0\linewidth}{!}
    {
    \label{tab:cls}
    \begin{tabular}{p{2.1cm} | p{1.5cm}<{\centering} p{1.5cm}<{\centering} p{2cm}<{\centering} p{1.5cm}<{\centering} | p{1.5cm}<{\centering} | p{1.5cm}<{\centering} | p{1.5cm}<{\centering}}
        \toprule
        \multirow{2}{*}[-0.6ex]{Methods} & \multirow{2}{*}[-0.6ex]{Params (M)} & \multirow{2}{*}[-0.6ex]{FLOPs (G)} & \multirow{2}{*}[-0.6ex]{Throughput
(img/s)} & \multirow{2}{*}[-0.6ex]{Mem (MB)} & \multicolumn{3}{c}{Top-1 Acc (\%)}\\
        \cmidrule{6-8}
        & & & & &ISIC-2018  & ColonPath & Brain Tumor\\
        \midrule
        HiFuse-Tiny\cite{HUO2024105534} & 120.00 & 17.56 & 415.6 & 5599 & 84.61 & \underline{98.70} & 95.14\\
        MIL-VT\cite{yu2021mil-vt} & 21.80 & 4.28 & 1288.0 & 1027 & 84.21 & 97.55 & 87.13\\
        KAT \cite{Zheng2023KAT}  & 20.08 & 3.01 & 879.8 & 1975 & 88.54 & 98.22 & 96.31\\
        MG-Trans \cite{Shi2023MG-Trans} & 22.12 & 3.38 & 926.7 & 2033 & 88.32 & 98.46 & \underline{97.16}\\
        \midrule
       ViT-S\cite{vit} & 21.67 & 4.61 & 1299.5 & 1027 & 87.75 & 97.30  & 88.21\\
       PVT-S\cite{wang2021PVT} & 23.98 & 3.83 & 917.0 & 1985 & \underline{88.76} & 97.30 & 90.52\\
       Swin-T\cite{liu2021swin} & 27.53 & 4.51 & 802.8 & 2323 & 87.72 & 98.10 & 91.43\\
       BiFormer-S\cite{zhu2023biformer} & 25.03 & 4.49 & 996.6 & 2091 & 86.22 & 98.35 & 96.33\\
       \midrule
       MedFormer-T & 3.20 & 0.58  & 1569.3 & 1715 & 85.51 & 97.29 & 91.57\\
       MedFormer-S & 12.63 & 2.18 & 941.2 & 2215 & 88.24 & 98.10 & 95.62\\
       MedFormer-B & 28.30 & 4.79 & 600.3 & 2993 & \textbf{89.71} & \textbf{98.72} & \textbf{98.13}\\
       \midrule
       MedFormer-T* & 3.20 & 0.58  & 1569.3 & 1715 & 93.61 & 100.00 & 99.64\\
       MedFormer-S* & 12.63 & 2.18 & 941.2 & 2215 & 94.44  & 100.00 & 99.83\\
       MedFormer-B* & 28.30 & 4.79 & 600.3 & 2993 & 94.57 & 100.00 & 99.91\\
        \bottomrule
    \end{tabular}
    }
\end{table*}

\subsection{Architecture Design}
\subsubsection{MedFormer Block}
Utilizing DSSA, we construct a basic block called the MedFormer Block. This block is composed of three main components: a 3$\times$3 depth-wise convolution to implicitly encode relative positional information, a DSSA layer performs sparse selection both at region-level and pixel-level to capture critical features, and a two-layer MLP module with expansion ratio $e$=3. Each component incorporates residual connections to maintain the model’s learning capacity, and layer normalization is applied to stabilize the training process and improve the capability of generalization. The detailed design of the block is shown in Fig. \ref{fig3:backbone}.

\subsubsection{Architecture}
To produce hierarchical feature maps for enhancing performance in classification and various dense prediction tasks, we construct a hierarchical pyramid medical backbone network based on MedFormer Blocks, named MedFormer. The detailed design is illustrated in Fig. \ref{fig3:backbone}. 
Specifically, Stage 1 consists of a patch embedding layer and $N_1$ MedFormer Blocks. The patch embedding layer is composed of two 3$\times$3 convolutions, which transform the input image into a feature map with a resolution of $\frac{H}{4} \times \frac{W}{4} \times C$. To obtain a hierarchical representation, Stage 2 introduces a patch merging layer with a 3$\times$3 convolution, which reduces the number of tokens and sets the channel dimension to $2C$, resulting in a feature map size of $\frac{H}{8} \times \frac{W}{8} \times 2C$. Then, $N_2$ MedFormer Blocks are applied to extract effective features. After repeating the process twice, Stage 3 and Stage 4 generate feature maps with dimensions $\frac{H}{16} \times \frac{W}{16} \times 4C$ and $\frac{H}{32} \times \frac{W}{32} \times 8C$, respectively. Additionally, by adjusting the parameters of MedFormer, we proposed three different variants: Tiny, Small and Base. Table \ref{tab1:version} presents the detailed parameter configurations of these MedFormer variants for classification tasks.

\section{Experiments and Results}
\label{sec:exp-results}
This section begins with an overview of the datasets used in our study. Next, we evaluate the Performance of MedFormer across three tasks: classification, semantic segmentation, and object detection, detailing the corresponding experimental setups and results. Then, we conduct a series of ablation studies to validate the contributions of each component of MedFormer, including the effectiveness of DSSA, hyperparameter selection, the effectiveness of denoising, and the effect of model depth on performance. Finally, we provide comprehensive visualization analyses. All experiments are conducted using the PyTorch 2.0 framework on four NVIDIA 3090 graphics cards with 24GB memory.

\subsection{Datasets}
This subsection introduces the datasets used for each task, with detailed information about each dataset provided in Table \ref{tab2:data}. Notably, for classification and detection datasets, we split them into training and test sets in an 8:2 ratio. As for segmentation datasets, we follow the same data partitioning scheme as previous works, such as Swin-Unet, MISSFormer, and BRAU-Net++.

\subsubsection{ISIC-2018 Classification}
The classification dataset originates from the ISIC-2018 Challenge Task 3 and encompasses seven distinct categories of images. All images have been standardized to a uniform resolution of 600$\times$450 pixels. The entire dataset consists of 9,376 training images and 2,344 test images.
\subsubsection{ColonPath}
This is a binary classification dataset from Ruijin Hospital in China, designed to detect malignant cell regions in tissue slides. These images are derived from colonoscopy results of 396 patients. We randomly split the dataset into 8,007 training images and 2,002 test images.
\subsubsection{Brain Tumor}This is an MRI image dataset for four-class brain tumor classification, consisting of a total of 7,023 images, including 5,618 training images and 1,405 test images.
\subsubsection{ISIC-2018 Segmentation}
This segmentation dataset is collocated from  the ISIC-2018 Challenge Task 1 and contains 2,594 images with varying resolutions. 
Following \cite{Lan2024BRAU-Net++}, we first allocate 10\% as the test set, then split the
remaining images into training and validation sets with 8:2 ratio.
\subsubsection{CVC-ClinicDB}
CVC-ClinicDB is a widely used polyp segmentation dataset consisting of 612 images. Following the same division method as the ISIC-2018 Segmentation dataset, we obtain 441 training images, 110 validation images, and 61 test images.
\subsubsection{Synapse}
This is a multi-organ segmentation dataset consists of 30 abdominal CT scans, with each CT volume containing between 85 and 198 slices of 512$\times$512 pixels. The dataset is partitioned following \cite{cao2022swin_unet}, \cite{huang2022missformer}.  Specifically, 2,212 slices from 18 scans are used for training, and 1,567 slices from 12 scans are used for testing.
\subsubsection{Kvasir-Seg}
The Kvasir-Seg dataset contains 1,000 pathological polyp images and their corresponding annotations. We randomly split the dataset into 800 training images and 200 testing images.
\subsubsection{Brain Tumor Detection}
This dataset is specifically designed for detecting brain tumors using computer vision techniques. The train set and test set includes 7,128 images and 1,782 images, respectively.

\begin{table*}
\centering
\small
\caption{Quantitative results on Params, FLOPs, DSC, and HD of our approach against other state-of-the-art methods for medical image segmentation on Synapse multi-organ segmentation dataset. Only DSC is exclusively used for the evaluation of individual organ. The symbol $\uparrow$ indicates the larger the better. The symbol $\downarrow$ indicates the smaller the better. The best result is in \textbf{Blod}, and the second best is \underline{underlined}. * indicates the model is initialized with pre-trained weights.}
\resizebox{1.0\linewidth}{!}{
\begin{tabular}{p{2.6cm} | p{1.3cm}<{\centering} p{1.3cm}<{\centering} | p{1.25cm}<{\centering} p{1.3cm}<{\centering} | p{0.8cm}<{\centering} p{1.35cm}<{\centering} p{1.2cm}<{\centering} p{1.2cm}<{\centering} p{0.8cm}<{\centering} p{1.0cm}<{\centering} p{0.9cm}<{\centering} p{1.0cm}<{\centering}}
\toprule
Methods & Params~(M)& FLOPs~(G)  & DSC~(\%)~$\uparrow$  & HD~(mm)~$\downarrow$ & Aorta & Gallbladder & Kidney~(L) &Kidney~(R) &Liver & Pancreas & Spleen & Stomach\\
\midrule 
Focal-UNet \cite{naderi2022focal_unet} & 44.25 & 9.99 &80.81  &20.66& 85.74 & 71.37  &\underline{85.23} &\underline{82.99}  &94.38 & 59.34 &88.49 & 78.94
\\
MISSFormer \cite{huang2022missformer} &42.46 & 9.89 & \underline{81.96}  & \underline{18.20} & 86.99 &68.65  &85.21& 82.00 &94.41 & \underline{65.67} &91.92 & 80.81
\\
MAXFormer \cite{liang2023maxformer}& 88.93 & 45.16 & \textbf{83.66} & \textbf{15.89} & \textbf{87.72} & \textbf{73.53} & \textbf{87.92} & \textbf{84.67} & \underline{95.00} & \textbf{66.55} & \underline{92.46} & \underline{81.44}\\
MedFormer-B & 31.28 & 7.49 & 81.75 & 18.72 & \underline{87.62} & \underline{71.66} & 82.25 & 81.04 & \textbf{95.14} & 62.12 & \textbf{92.52} & \textbf{81.65}\\
\midrule
\midrule
TransUNet* \cite{chen2021transunet} &105.28 & 29.35 & 77.48   &31.69 &87.23 &63.13 &81.87 &77.02 &94.08 &55.86 &85.08 &75.62 
\\
Swin-Unet* \cite{cao2022swin_unet} & 27.17 & 6.16 & 79.13  & 21.55 & 85.47 & 66.53  & 83.28 & 79.61  & 94.29 & 56.58 & 90.66  & 76.60
\\
HiFormer* \cite{heidari2023hiformer} &25.51 & 8.05 & 80.39  &\textbf{14.70} & 86.21 &65.69  & 85.23 &79.77  &94.61 & 59.52 &90.99 & 81.08
\\
BRAU-Net++* \cite{Lan2024BRAU-Net++} & 62.63 & 17.66 & 82.47  &19.07 & \underline{87.95} &69.10  &\textbf{87.13} &81.53 & 94.71  & 65.17 &91.89 & 82.26
\\
\midrule
PVT-S*\cite{pvt_cascade} & 25.35 & 4.49 & 80.00 & 23.47 & 85.87 & 69.28 & 80.09 & 77.44 & 94.77 & 60.25 & 90.65 & 81.67\\
Swin-T*\cite{liu2021swin} & 28.61 & 5.20 & 82.02 & 19.87 & 87.03 &	73.87 & 83.38 & 81.20 & 94.46 & 65.00 & 89.58 & 81.64\\
DAT-T*\cite{Xia2022deformable_attention} & 24.31 & 4.87 & 82.68 & 19.45 & 87.37 & 74.50 & 83.30 & 79.55 & 94.85 & 65.07  & 92.13 & \underline{84.69}\\
BiFormer-S*\cite{zhu2023biformer} & 25.89 & 5.55 & \underline{82.99} & \underline{17.15} & 87.17 & 71.22 & 85.87 & \underline{82.79} & \underline{94.94} & \textbf{67.42} & 91.02 & 83.45\\
\midrule 
MedFormer-T* & 3.83 & 1.26 & 81.23 & 20.20 & 85.93 & \underline{74.51} & 80.74 & 78.86 & 94.61 & 61.79 & 91.53 & 80.90\\
MedFormer-S* & 13.49 & 2.90 & 82.21 & 19.23 & 86.84 & 69.61 & 84.27 & 81.91 & 94.44 & 64.89 & \textbf{93.10} & 82.65\\
MedFormer-B* & 31.28 & 7.49 & \textbf{84.07} & 17.38 & \textbf{88.16} & \textbf{74.83} & \underline{85.90} & \textbf{82.90} & \textbf{95.58} & \underline{66.96} & \underline{93.01} & \textbf{85.24}\\
\bottomrule
\end{tabular} }
\label{tab4:synapse}
\end{table*}

\begin{table*}[htpb]
\centering
\caption{Quantitative results of different methods for medical image segmentation on ISIC-2018 Segmentation and CVC-ClinicDB datasets. * indicates the model is initialized with pre-trained weights.}
\resizebox{1.0\linewidth}{!}
{
    \begin{tabular}{p{2.3cm} | p{0.8cm}<{\centering} p{0.8cm}<{\centering} p{0.8cm}<{\centering} p{0.8cm}<{\centering} p{0.8cm}<{\centering}| p{0.8cm}<{\centering} p{0.8cm}<{\centering} p{0.8cm}<{\centering} p{0.8cm}<{\centering} p{0.8cm}<{\centering} | p{1.1cm}<{\centering} p{1.1cm}<{\centering}}
        \toprule
        \multirow{2}{*}[-0.6ex]{Methods} & \multicolumn{5}{c|}{ISIC-2018 Segmentation} & \multicolumn{5}{c|}{CVC-ClinicDB} & \multirow{2}{*}[-0.6ex]{\makecell{Params~(M)}} & \multirow{2}{*}[-0.6ex]{\makecell{FLOPs~(G)}} \\
        \cmidrule{2-11}
          & mIoU & DSC & Acc & Prec & Recall & mIoU & DSC & Acc & Prec & Recall \\
        \midrule
       MedT \cite{medt} & 81.43 & 86.92 & \textbf{95.1}0 & 90.56 & \textbf{89.93} & 81.47 & 86.97 & 98.44 & 89.35 & \textbf{90.04} & 1.56 & 3.60 \\
       MISSFormer \cite{huang2022missformer} & \underline{81.45} & \underline{88.29} & \underline{94.81} & \textbf{91.37} & 89.13 & \underline{83.68} & \underline{89.50} & \underline{98.58} & \textbf{91.65} & 88.08 & 42.53 & 11.92\\

       MedFormer-B & \textbf{81.67} & \textbf{88.42} & 94.78 & \underline{91.15} & \underline{89.60} & \textbf{84.4}9 & \textbf{89.99} & \textbf{98.71} & \underline{91.13} & \underline{89.55} & 29.38 & 7.15\\
       \midrule
       \midrule
       TransUNet* \cite{chen2021transunet} & 77.05 & 84.97 & 94.56 & 84.77 & 89.85 & 79.95 & 86.70 & 98.25 & 87.63 & 87.34 & 105.32 & 38.54\\
       Swin-Unet* \cite{cao2022swin_unet} & 81.87 & 87.43 & \underline{95.44} & 90.97 & \underline{91.28} & 84.85 & 88.21 & 98.72 & 90.52 & 91.13 & 27.18 & 8.09 \\
       HiFormer* \cite{heidari2023hiformer} & 81.54 & 88.44 & 94.68 & 90.23 & 90.82 & 85.74 & 90.48 & 98.62 & 91.56 & 90.37 & 25.51 & 8.05\\ 
       BRAU-Net++* \cite{Lan2024BRAU-Net++} & \textbf{84.01} & \textbf{90.10} & \textbf{95.61} & 91.18 & \textbf{92.24} & 88.17 & 92.94 & 98.83 & 93.84 & 93.06 & 62.63 & 23.03\\
       \midrule
       PVT-S* \cite{wang2021PVT} & 82.22 & 88.75 & 94.82 & 91.46 & 90.24 & 86.08 & 91.53 & 98.80 & 92.77 & 91.00 & 25.47 & 5.92\\
       Swin-T* \cite{liu2021swin} & 82.83 & 89.24 & 95.14 & 91.45 & 90.83 & 88.70 & 93.75 & 98.97 & 94.84 & 93.16 & 28.61 & 8.26\\
       DAT-T* \cite{Xia2022deformable_attention} & 83.07 & 89.43 & 95.25 & 91.43 & 91.17 & 90.21 & 94.33 & 99.01 & 95.43 & 94.31 & 24.31 & 6.40\\
       BiFormer-S* \cite{zhu2023biformer} & 83.17 & 89.54 & 95.23 & 91.32 & 91.13 & \underline{90.27} & \underline{94.69} & \underline{99.02} & \underline{95.47} & 93.97 & 25.89 & 6.74\\
       \midrule
       MedFormer-T* & \underline{83.26} & \underline{89.63} & 95.31 & \textbf{91.65} & 91.22 & 86.00 & 91.13 & 98.71 & 91.56 & 92.08 & 3.83 & 1.60\\
       MedFormer-S* & 82.69 & 89.20 & 95.11 & 91.35 & 90.95 & 89.22 & 94.02 & \underline{99.02} & 94.35 & \textbf{94.33} & 13.49 & 3.72\\
       MedFormer-B* & 82.96 & 89.45 & 95.34 & \underline{91.53} & 91.01 & \textbf{90.33} & \textbf{94.73} & \textbf{99.08} & \textbf{95.64} & \underline{94.28} & 29.38 & 7.15\\
        \bottomrule
    \end{tabular}
}
\label{tab5:isic & cvc}
\end{table*}

\subsection{Medical Image Classification}
\subsubsection{Settings}
The classification performance of MedFormer is evaluated on three medical image datasets. To ensure a fair and comprehensive evaluation, MedFormer is compared with two types of methods: specialized medical image classification methods and widely-used Transformer backbone networks. The AdamW optimizer is used with a weight decay of 0.05 and the initial learning rate is set to $1e-6$. A linear warm-up strategy is employed to gradually increase the learning rate during the first 5 epochs, followed by a cosine annealing schedule with a minimum learning rate of $1e-5$. To further enhance the model's generalization ability, a series of data augmentation techniques are applied, including mixup, cutmix, and random erasing. Additionally, the input images are resized to 224$\times$224, and the training is conducted for 300 epochs. For MedFormer, key parameters are configured, such as setting the region partition factor $S$ to 7, the hyperparameter $\lambda$ to 1/8, the dimension of each attention head to 32, and the MLP expansion ratio to 3.

\begin{table*}[t]
\centering
\caption{Results of different methods for Medical Image Detection on Kvasir-Seg and Brain Tumor Detection dataset. * indicates the model is initialized with pre-trained weights.}
\resizebox{1.0\linewidth}{!}
{
    \begin{tabular}{p{3cm} | p{0.7cm}<{\centering} p{0.7cm}<{\centering} p{0.7cm}<{\centering} p{0.7cm}<{\centering} | p{1.2cm}<{\centering} p{1.2cm}<{\centering} | p{0.7cm}<{\centering} p{0.7cm}<{\centering} p{0.7cm}<{\centering} p{0.7cm}<{\centering} | p{1.2cm}<{\centering} p{1.2cm}<{\centering}}
        \toprule
        \multirow{2}{*}[-0.6ex]{Methods} & \multicolumn{6}{c|}{Kvasir-Seg} & \multicolumn{6}{c}{Brain Tumor Detection}\\
        \cmidrule{2-13}
         & $mAP$ & $AP_{50}$ & $AP_{75}$ & Recall & Params~(M) & FLOPs~(G) & $mAP$ & $AP_{50}$ & $AP_{75}$ & Recall & Params~(M) & FLOPs~(G)\\
        \midrule
      RetinaNet* \cite{retinanet} & 51.2 & 87.5 & 57.4 & 91.9 & 36.33 & 13.05 & 35.8 & 70.1 & 32.5 & 82.3 & 36.37 & 13.11\\
      Faster R-CNN* \cite{faster_rcnn} & 48.7 & 85.3 & 57.5 & 90.7 & 41.35 & 26.21 & 38.0 & 72.3 & 35.7 & 77.7 & 41.36 & 26.23\\
       Deformable DETR* \cite{zhu2020deformable_detr} & 54.3 & 86.0 & \underline{62.4} & \textbf{96.1} & 41.07 & 15.21 & 38.7 & 73.6 & 37.3 & \textbf{87.2} & 41.08 & 15.21\\
       \midrule
       PVT-S*\cite{wang2021PVT} & \underline{56.2} & 84.3 & 62.2 & 89.9 & 32.58 & 12.51 & \underline{45.6} & 76.6 & \underline{47.8} & 83.6 & 32.62 & 12.56\\
       Swin-T*\cite{liu2021swin} & 55.8 & 89.0 & 62.3 & 94.9 & 36.82 & 14.58 & 41.9 & 75.2 & 41.3 & 85.9 & 36.86 & 14.64\\
       DAT-T*\cite{Xia2022deformable_attention} & 47.3 & 85.3 & 49.5 & 92.4 & 32.05 & 12.83 & 41.5 & 74.3 & 41.3 & 85.8 & 32.09 & 12.89\\
       BiFormer-S*\cite{zhu2023biformer} & 49.4 & 88.6 & 48.7 & 95.3 & 33.62 & 13.32 & 45.5 & \underline{76.8} & \underline{47.8} & 85.6 & 33.66 & 13.38\\
       \midrule
       MedFormer-T* & 52.6 & 88.8 & 58.1 & \underline{95.3} & 11.09 & 8.17 & 42.6 & 74.6 & 43.5 & \underline{86.2} & 11.13 & 8.23\\
       MedFormer-S* & 53.8 & \underline{89.3} & 60.6 & 93.0 & 21.22 & 10.29 & 43.9 & 76.1 & 45.2 & 84.9 & 21.27 & 10.35\\
       MedFormer-B* & \textbf{56.4} & \textbf{90.1} & \textbf{62.6} & 94.6 & 37.59 & 13.34 & \textbf{45.8} & \textbf{77.1} & \textbf{48.1} & 84.2 & 37.63 & 13.80\\
       \bottomrule
    \end{tabular}
}
\label{tab6:detection}
\end{table*}

\subsubsection{Results}
All classification results are summarized in Table \ref{tab:cls}. It is observed that MedFormer-B achieves the best performance across all three classification datasets. Notably, MedFormer-T achieves comparable performance to some methods with only 3.2M parameters and 0.58G FLOPs. In contrast, HiFuse-Tiny requires significantly more resources but performs well only on the ColonPath dataset. KAT and MG-Trans are effective in reducing computational requirements, but their use of kernel techniques or graph transformers may overlook the extraction of detailed features, thus limiting the improvement in accuracy. The accuracy of MedFormer surpasses that of the compared backbone networks, demonstrating MedFormer's superior feature representation ability. Additionally, MedFormer with pre-trained weights achieves an accuracy of 100\% and 99.91\% on the ColonPath and Brain Tumor datasets, respectively, showcasing its immense potential in medical image classification tasks. 

It is worth noting that the pretrained version of MedFormer is included solely to demonstrate its potential but is not incorporated into the formal comparisons.
To further evaluate the deployment feasibility of these methods in real clinical systems, we measure the throughput and memory usage for each method. The results are shown in Table \ref{tab:cls}. It can be seen that ViT-S and MIL-VT, although capable of processing images at a high throughput with relatively low memory usage, exhibit lower accuracy. MedFormer-T is suitable for scenarios where speed is prioritized, offering good throughput and low memory usage. MedFormer-B, though slower, excels in accuracy and is ideal for medical image classification tasks requiring high precision. MedFormer-S strikes a good balance between efficiency and accuracy, making it suitable for most clinical applications. 
These variants ensure that MedFormer can effectively meet the needs of various clinical scenarios.

\subsection{Medical Image Segmentation}
\subsubsection{Settings}
MedFormer is employed as the backbone network in UPerNet \cite{xiao2018unified}, and its performance is evaluated against other popular backbone networks on medical image segmentation tasks.
Furthermore, comparisons are conducted between MedFormer and several state-of-the-art (SOTA) medical segmentation methods to emphasize its competitiveness. For the ISIC-2018 Segmentation and CVC-ClinicDB datasets, the input image size is set to 256$\times$256. The Adam optimizer is used with an initial learning rate of $1e-4$, and training runs for 200 epochs with a cosine annealing learning rate schedule. For the Synapse dataset, the input size is adjusted to 224$\times$224, the AdamW optimizer is utilized, and training runs for 400 epochs using a cosine annealing learning rate schedule with an initial learning rate of $1e-3$. To avoid overfitting, data augmentation techniques such as horizontal flipping, vertical flipping, and scaling and rotation are applied. For the key parameters of MedFormer, the dropout rate is set to 0.2, while the region partition factor $S$ is 8 for a resolution of 256$\times$256 and 7 for 224$\times$224.

\subsubsection{Results}
Table \ref{tab4:synapse} presents the experimental results on the Synapse dataset. In methods without pretraining, MAXFormer achieves leading results, even surpassing some methods that use pretrained weight initialization. In methods with pretrained weights, MedFormer-B* achieves the best average DSC of 84.07\%, outperforming all other comparison methods. The result indicates the significant advantage of MedFormer in accurately overlapping the predicted results with the Ground Truth. However, MedFormer does not achieve the highest score in the HD metric, showing a deficiency in learning edge information. We hypothesize that this issue arises because DSSA performs twice sparse selections to focus on a small number of important tokens. Since the number of boundary tokens is much smaller than that of internal organ tokens, and the boundary tokens often have semantic ambiguity, they are overlooked during the selection process. This limitation becomes more evident in multi-organ segmentation datasets with blurry boundaries. Additionally, We observe that while MedFormer-B* achieves the best DSC (84.07\%) in the segmentation task, the performance improvement over MAXFormer (83.66\%) is relatively small. To verify whether this difference is statistically significant, we performed a paired t-test to compare the performance of the two models on the same dataset. After conducting statistical analysis on the Dice coefficient differences for each sample, we obtain a p-value of 0.048, which is below the common significance level of 0.05. This result indicates that the performance improvement is statistically significant, further supporting the credibility and effectiveness of the proposed method.
Furthermore, the segmentation results on the ISIC-2018 Segmentation and CVC-ClinicDB dataset are shown in Table \ref{tab5:isic & cvc}. 
Without using pre-trained weights, MedFormer-B achieves performance comparable to SOTA methods on both datasets. In the comparison of pre-training initialization strategies, MedFormer-B* achieves leading results on the CVC-ClinicDB dataset. Although MedFormer outperforms most competing methods on the ISIC-2018 segmentation dataset, there remains a slight performance gap compared to BRAU-Net++. We consider this result understandable for two reasons: first, the marginal advantage of BRAU-Net++ comes at the cost of significantly higher computational complexity; second, MedFormer is not a dedicated segmentation model and only employs a simple segmentation head—UPerNet. Under the same UPerNet configuration, other backbone networks perform worse than MedFormer. Overall, MedFormer demonstrates robust performance in segmentation tasks, highlighting its potential as a general-purpose backbone network.

\subsection{Medical Image Detection}
\subsubsection{Settings}
Since Transformer-based medical image detection methods with available code are scarce, we reproducing several milestone object detection approaches to highlight the competitiveness of MedFormer in detection tasks. Furthermore, we also compare MedFormer with several popular backbone networks under the RetinaNet\cite{retinanet} detection framework to verify its effectiveness. The input resolution is set to 256$\times$256, and the AdamW optimizer is employed with a cosine annealing learning rate scheduler and a weight decay of $1e-4$. The initial learning rate is set to $1e-4$. The parameters for MedFormer are adjusted as follows: the region partition factor $S$ is set to 8, and the drop rate is set to 0.3.
\subsubsection{Results}
We use four metrics to evaluate the detection performance of all methods: mean Average Precision ($mAP$), $AP_{50}$, $AP_{75}$, and Recall. The detailed results are presented in Table \ref{tab6:detection}.
MedFormer-B achieves the best results on the Kvasir-Seg and Brain Tumor Detection datasets in terms of three metrics: $mAP$, $AP_{50}$ and $AP_{75}$. This proves the strong object detection performance of MedFormer. Furthermore, we observe that the differences in detection results among the comparative methods on the Kvasir-Seg dataset are not significant. However, on the Brain Tumor Detection dataset, dedicated detection methods generally lag behind the backbone networks. We hypothesize that this phenomenon is due to two main reasons: First, the Brain Tumor Detection dataset is multi-class and has nested bounding boxes, which makes accurate localization challenging for the models. Second, the backbone networks involved in the comparison are carefully designed and have stronger feature representation capabilities. All backbone networks are experimented within the RetinaNet framework, replacing the original ResNet encoder, achieving significant improvements. This further confirms that the primary gap exists in the encoder part.

\begin{table}
\centering
\caption{Ablation study of different attention mechanisms, on the proposed MedFormer architecture.}
\label{tab:attention}
\resizebox{1.0\linewidth}{!}
{
    
    \begin{tabular}{p{2.5cm} | p{2cm}<{\centering} | p{2cm}<{\centering}}
        \toprule
        \multirow{2}{*}[-0.9ex]{Attention} & ISIC-2018 & Synapse\\
        \cmidrule{2-3}
        & Top-1 Acc. (\%) & DSC (\%)\\
        \midrule
        Shifted window\cite{liu2021swin} & 85.49 & 72.50\\
        Explicit Sparse\cite{zhao2019explicit} & 86.53 & 78.66 \\
        Deformable\cite{Xia2022deformable_attention} & 87.56 &  73.03\\
        Bi-level Routing\cite{zhu2023biformer} & \underline{87.65} & \underline{79.35}\\
        Dual Sparse Selection & \textbf{88.24} & \textbf{79.44}\\
        \bottomrule
    \end{tabular}
}
\end{table}
\begin{table}[h]
\centering
\caption{Ablation study on the effect of LCE module on both ISIC-2018 Classification and Synapse Segmentation datasets.}
\label{tab:lce}
\resizebox{\linewidth}{!}
{
    \begin{tabular}{p{3cm} | p{2cm}<{\centering} | p{2cm}<{\centering}}
        \toprule
        \multirow{2}{*}[-0.9ex]{Methods} & ISIC-2018 & Synapse\\
        \cmidrule{2-3}
        & Top-1 Acc. (\%) & DSC (\%)\\
        \midrule
        MedFormer-S & \textbf{88.24} & \textbf{79.44}\\
        MedFormer-S (w/o LCE)& 87.64 & 78.93\\
        \midrule
        \midrule
        MedFormer-B & \textbf{89.71} & \textbf{81.75}\\
        MedFormer-B (w/o LCE)& 88.43 & 80.54\\
        \bottomrule
    \end{tabular}
}
\end{table}
\begin{table}[h]
\centering
\caption{Ablation study of the number of $k_1$ and $\lambda$ on ISIC-2018 Classification and CVC-ClinicDB dataset.}
\label{tab8:k1&lambda}
\resizebox{1.0\linewidth}{!}
{
    
    \begin{tabular}{p{1cm}<{\centering} | p{0.5cm}<{\centering} | p{1.3cm} <{\centering} | p{1.2cm} <{\centering} | p{1.3cm} <{\centering} | p{1.2cm} <{\centering}}
        \toprule
        \multirow{2}{*}[-0.8ex]{$k_1$} & \multirow{2}{*}[-0.8ex]{$\lambda$} & \multicolumn{2}{c|}{ISIC-2018 Classification} & \multicolumn{2}{c}{CVC-ClinicDB}\\
        \cmidrule{3-6}
        & & FLOPs (G) & Acc (\%) & FLOPs (G) & DSC (\%) \\
        \midrule
        1,2,8,49 &  1/8 & 2.15 & 87.05 & 3.69 & 87.24\\
        \midrule
        \multirow{4}{*}{1,4,16,49} & 1 & 2.23 & 87.44 & 3.79 & 87.25\\
         & 1/4 & 2.18 & 87.56 & 3.73 & 87.64\\
         & 1/8 & 2.18 & \textbf{88.24} & 3.32 & \textbf{88.23}\\
         & 1/16 & 2.17 & 85.84 & 3.31& 86.87\\
         \midrule
         2,8,32,49 & 1/8 & 2.25 & 87.56 & 3.73 & 87.55\\
        \bottomrule
    \end{tabular}
}
\end{table}

\begin{table}
\centering
\caption{Ablation study of model depth on ISIC-2018 Classification dataset.}
\label{tab9:depth}
\resizebox{1.0\linewidth}{!}
{
       \begin{tabular}{p{1.7cm}<{\centering} | p{1.5cm}<{\centering} | p{1.5cm}<{\centering} | p{1.5cm}<{\centering}}
        \toprule
        Depth & Params (M) & FLOPs (G) & Acc (\%) \\
        \midrule
        2,2,6,2 & 11.298 & 1.906 & 86.01\\
        2,2,8,2 & 12.633 & 2.184 & \textbf{88.24}\\
        4,4,8,4 & 18.35 & 3.026 & 86.87\\
        4,4,18,4 & 25.027 & 4.375 & 86.87\\
        \bottomrule
    \end{tabular}
}
\end{table}

\begin{figure}
\centering
\includegraphics[width=1\linewidth, keepaspectratio]{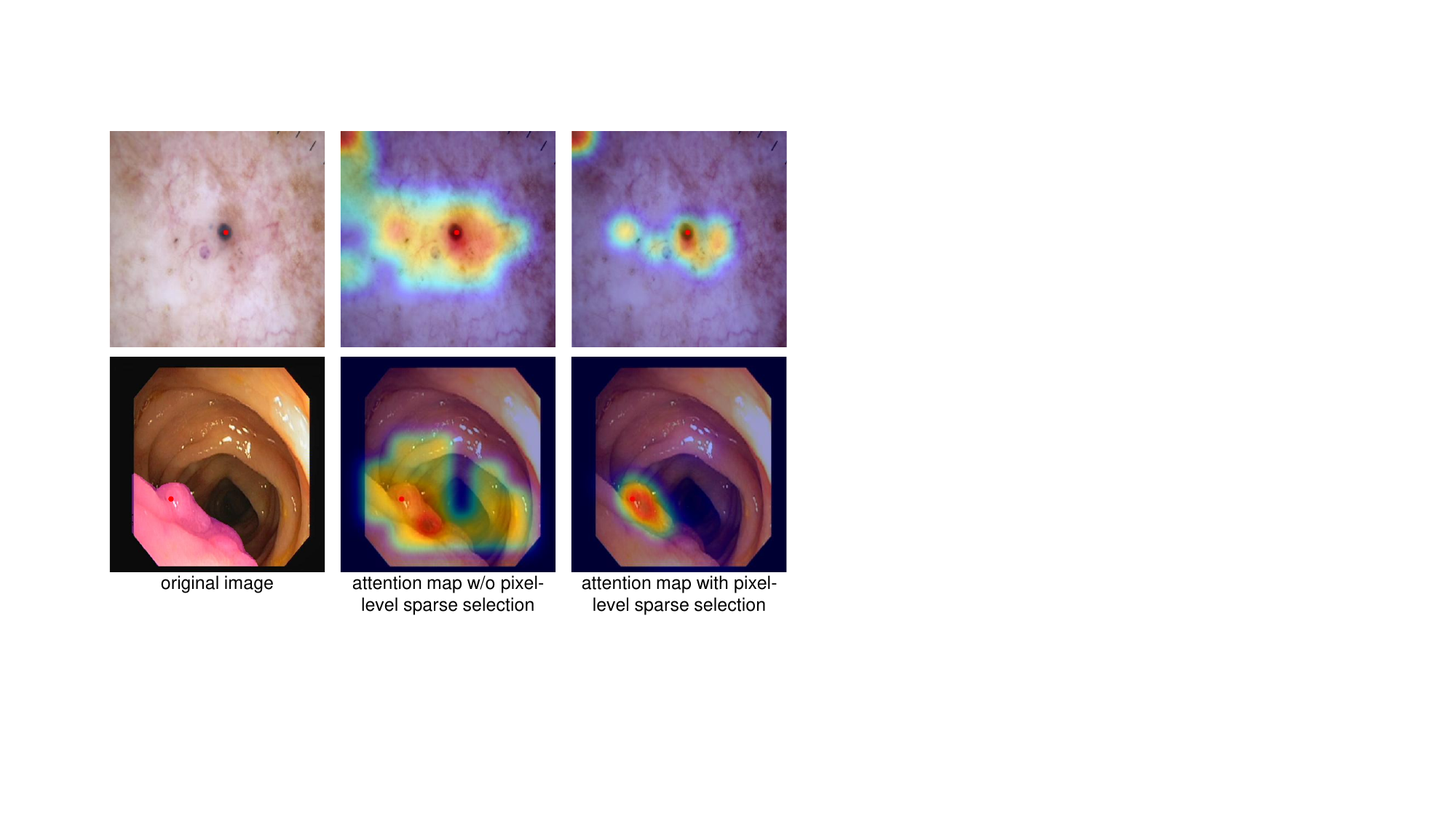}
\caption{Visualization of attention maps from sample images in the ISIC 2018 Classification dataset (top) and the CVC-ClinicDB segmentation dataset (bottom), illustrating the effects of excluding noise tokens. The pink regions in the segmented samples serve as the ground truth. The red circle represents the query token. It can be seen that pixel-level sparse selection enhances focus on lesion areas, which is particularly important for extracting critical features.}
\label{fig2:noise}
\end{figure}

\subsection{Ablation Study}
\subsubsection{The effectiveness of DSSA}
We compare existing sparse attention mechanisms on the ISIC-2018 Classification dataset and the Synapse segmentation dataset. Experiments are conducted under the MedFormer-S architecture. Table \ref{tab:attention} presents the results. Dual Sparse Selection attention achieves the best performance on both tasks. This can be attributed to its ability to exclude noise tokens while retaining global modeling capabilities, thus enabling the extraction of more  effective features. The Shifted Window attention performs poorly, likely due to the handcrafted sparse pattern failing to effectively model global dependencies. Deformable attention shows unsatisfactory results in multi-organ segmentation, as the key-value pairs are shared across all queries, which is disadvantageous for pixel-level prediction tasks. In contrast, the Explicit Sparse attention selects the most relevant key-value pairs for each query, leading to excellent segmentation performance. Leveraging its content-aware properties, the Bi-level Routing attention effectively select relevant tokens, which contributes to its strong performance on both tasks.

\subsubsection{The effectiveness of LCE}
We conduct ablation experiments on the LCE module across different MedFormer variants and tasks, with the results summarized in Table \ref{tab:lce}. As shown in the results, the LCE module provides a slight but consistent performance improvement across various tasks and model variants. This improvement can be attributed to the module's use of a 5$\times$5 depth-wise convolution, which effectively integrates local contextual information while remaining computationally lighter than standard convolutions.

\subsubsection{Choices of \texorpdfstring{$k_1$}{k1} and \texorpdfstring{$\lambda$}{lambda}}
In this work, we propose two key parameters, $k_1$ and $\lambda$, which control the sparse selection at the region-level and pixel-level, respectively. We gradually increase $k_1$ at each stage to ensure sufficient feature extraction in smaller regions. Table \ref{tab8:k1&lambda} presents the experimental results for the combination of these parameters. First, based on the analysis of the $k_1$ values, we observe that the number of tokens each query can attend to is limited, and introducing too many tokens actually leads to a degradation in performance. Next, we tested three different values of $\lambda$ (with the case of $\lambda = 1$ discussed in detail in the next section), and found that the intermediate value of $\lambda = 1/8$ yields the best performance on both datasets. We believe this result is due to the model requiring an appropriate number of pixel-level tokens to effectively learn features. Retaining too many pixel-level tokens may introduce excessive noise, distracting the model from key features, while retaining too few may result in insufficient information, limiting the model's generalization ability. Ultimately, the $k_1$ and $\lambda$ hyperparameters we selected achieve the best performance on both datasets, demonstrating their potential generalizability across tasks and datasets.

\begin{figure*}
\centering
\includegraphics[width=1\linewidth]{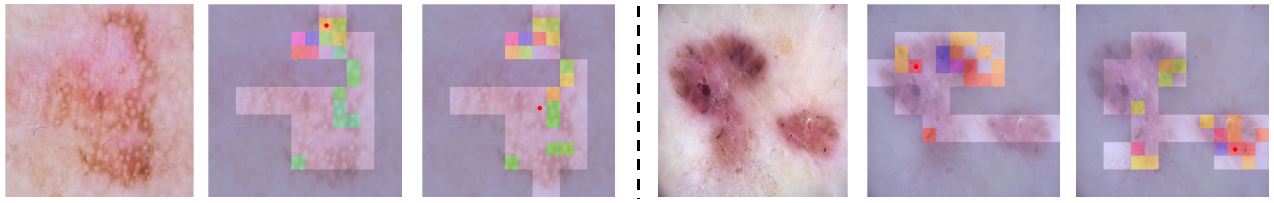}
\caption{Visualization of attention maps on ISIC-2018 Classification dataset. For each input, we select two different query positions to observe the region-level tokens they attend to and the pixel-level tokens within those regions. The heatmap values of the pixel-level tokens vary based on their relevance to the query. For this visualization, we use the region-level and pixel-level attention scores extracted from the final MedFormer block of the third stage.}
\label{fig:attention}
\end{figure*}

\begin{figure*}
\centering
\includegraphics[width=1\linewidth]{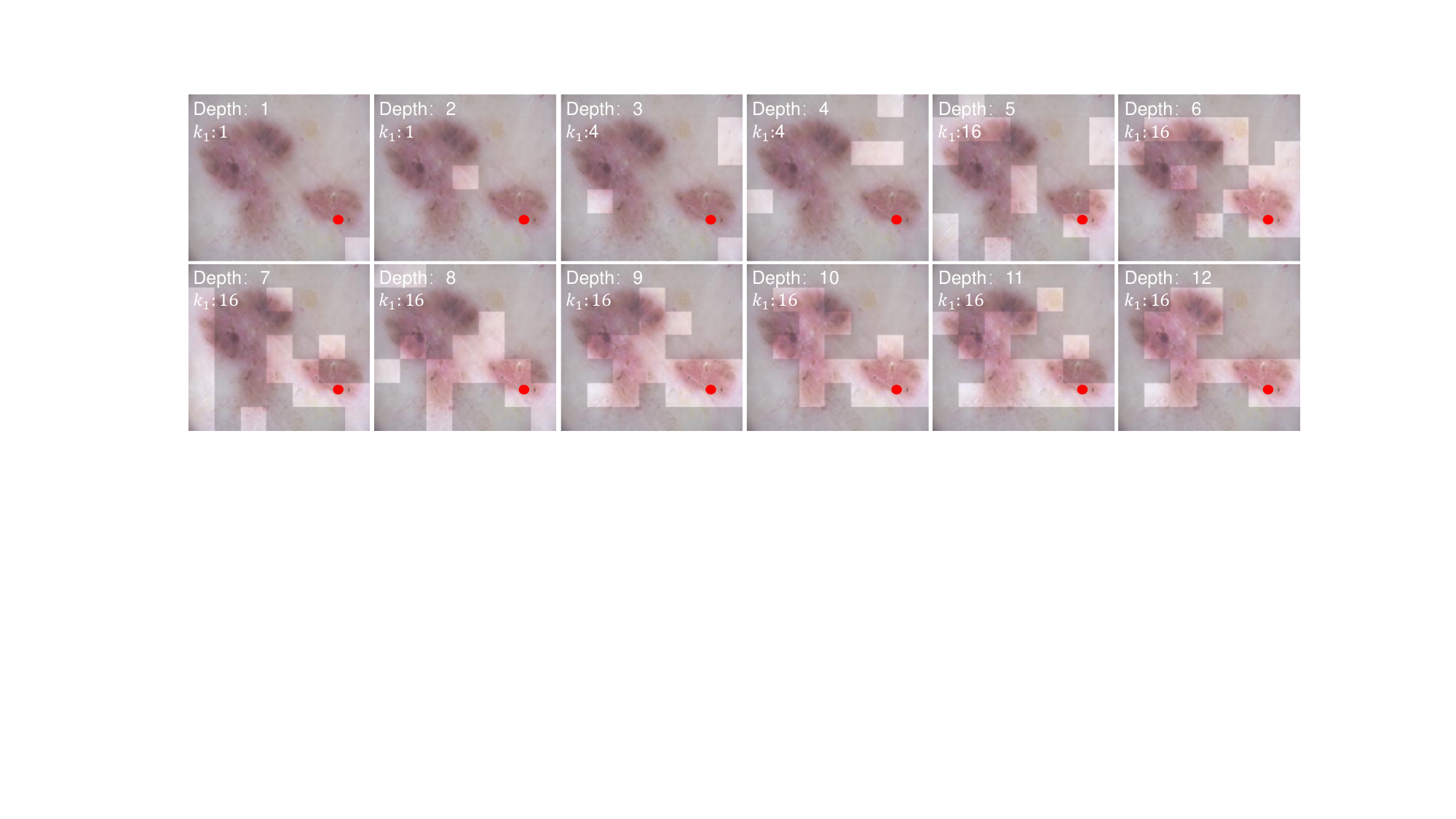}
\caption{On the ISIC-2018 Classification dataset, we visualize the regional attention maps at a fixed query position across network depths ranging from 1 to 12. The number of regions each query is determined by $k_1$.}
\label{fig:stage_attention}
\end{figure*}

\begin{figure*}
\centering
\includegraphics[width=1\linewidth]{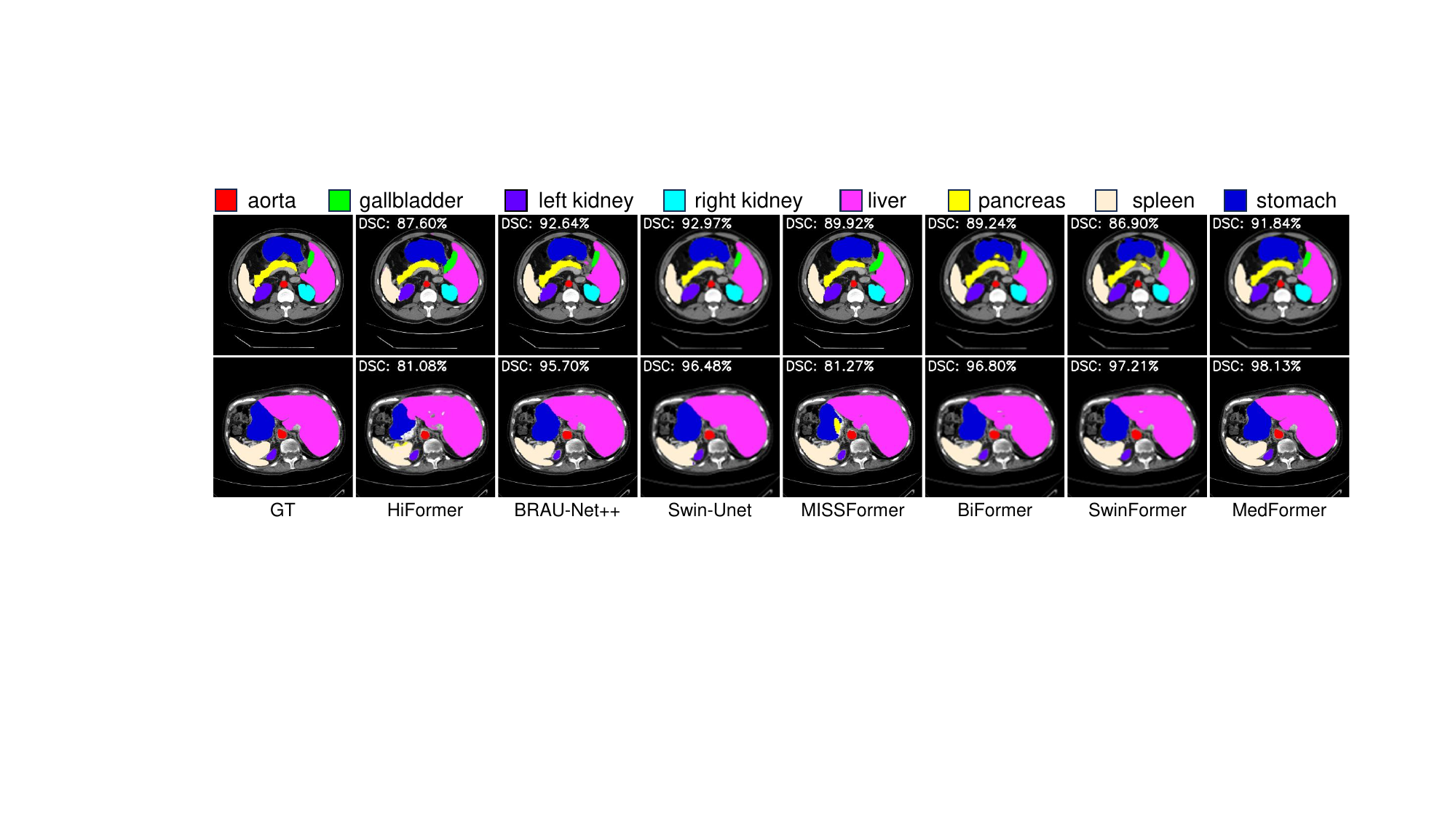}
\caption{Visualization of different methods on the Synapse dataset. MedFormer achieves finer segmentation results compared to other methods, showing a close alignment with the Ground Truth. In contrast, other approaches suffer from category confusion or fragmented segmentation regions.}
\label{fig:synapse}
\end{figure*}

\subsubsection{Effectiveness of Excluding Noise}
We design ablation experiments to confirm that pixel-level sparse selection can effectively exclude noise. First, we present quantitative results in Table \ref{tab8:k1&lambda}. The accuracy of the model significantly decrease  when the selection is not applied ($\lambda=1$). We attribute this drop to the introduction of excessive noise tokens, which prevent the model from extracting relevant features. Second, we visualize the attention map with and without pixel-level sparse selection in Fig. \ref{fig2:noise}. The results clearly show that excluding noise tokens allows the model to focus more on the lesion area. Furthermore, from the segmentation samples, we observed that the attention maps before and after sparse selection exhibit different sensitivities to the boundaries. After sparse selection, the attention converges significantly to the interior of the ground truth boundaries. This suggests that DSSA can, to some extent, adaptively align with the biological boundary features of the lesions.

\subsubsection{Effect of the depth}
We explore the impact of model depth on performance in MedFormer-Small, as shown in Table \ref{tab9:depth}. We find that an appropriate depth helps to effectively extract and aggregate semantic information, thereby improving model performance. However, continuously stacking MedFormer blocks not only significantly increases the parameter count and FLOPs but may also lead to a decrease in model accuracy. To balance best performance and computational efficiency, we set the depth to [2,2,8,2].

\subsection{Visualization Study}
In this section, we will conduct a series of visualization studies: visualization of attention maps, visualization of MedFormer on segmentation datasets, and visualization of MedFormer on object detection datasets.
\subsubsection{Visualization of Attention Map}
To intuitively demonstrate the superiority of the proposed DSSA, we visualize the attention maps of DSSA at both the region and pixel levels, as shown in Fig. \ref{fig:attention}. We observe that DSSA first selects the most relevant regions and then focuses on the tokens most related to the query within those regions. This aligns well with our design philosophy and validates the rationality and effectiveness of our approach.
In addition, we visualize the regional attention changes across each layer in MedFormer, as shown in Fig. \ref{fig:stage_attention}. This visualization shows that as the number of layers increases, our method progressively focuses on relevant regions, demonstrating its ability to refine attention dynamically and align with clinically meaningful areas.

\subsubsection{Visualization of Semantic Segmentation}
\begin{figure*}
\centering
\includegraphics[width=1\linewidth]{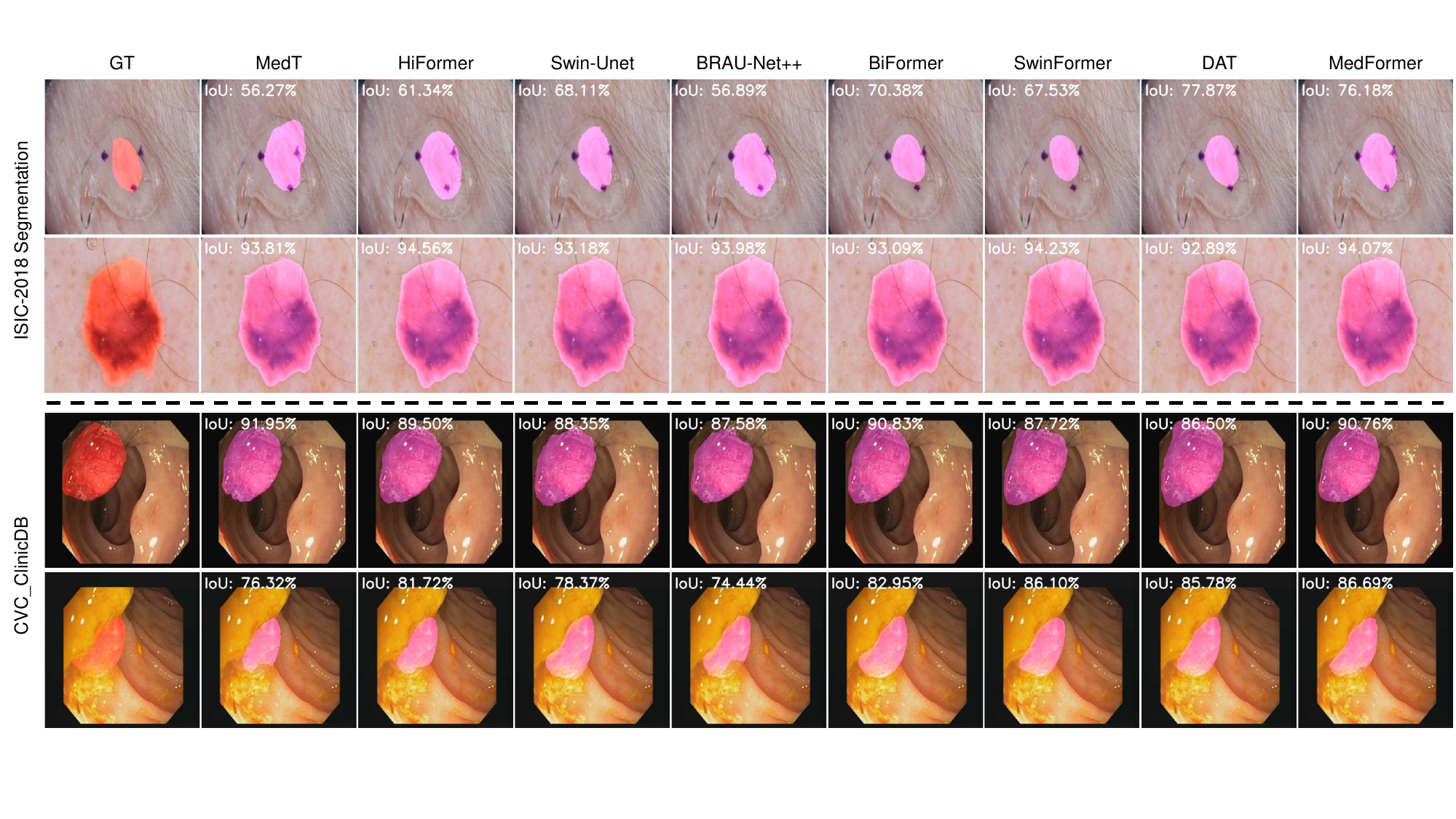}
\caption{Visualizations of Segmentation using various methods on the CVC-ClinicDB and ISIC-2018 Segmentation datasets. The lesion regions of GT are shown in \textcolor{red}{red}, while the segmentation results of methods are shown in \textcolor{pink}{pink}.}
\label{fig:isic_cvc}
\end{figure*}

We visualize the segmentation results of MedFormer, SOTA medical image segmentation methods and popular backbone networks on three datasets. Fig. \ref{fig:synapse} shows the results on the Synapse dataset, where MedFormer’s segmentation results closely matches the ground truth. In contrast, other methods exhibit issues such as class misidentification and loss of internal details. Fig. \ref{fig:isic_cvc} presents the results on the ISIC-2018 Segmentation and CVC-ClinicDB datasets. For samples with clear boundaries (Sample 1 and Sample 3), most methods achieve competitive segmentation results. However, for cases with blurred edges or overexposure (Sample 2 and Sample 4), all methods exhibit significant performance degradation and high variability. Notably, MedFormer still maintains leading performance under these challenging conditions, demonstrating strong robustness in complex scenarios. Overall, MedFormer demonstrates superior performance in segmentation tasks due to its unique dual sparse selection attention mechanism, which enables precise extraction of detailed features.

\subsubsection{Visualization of Object Detection}
\begin{figure*}
\centering
\includegraphics[width=1\linewidth]{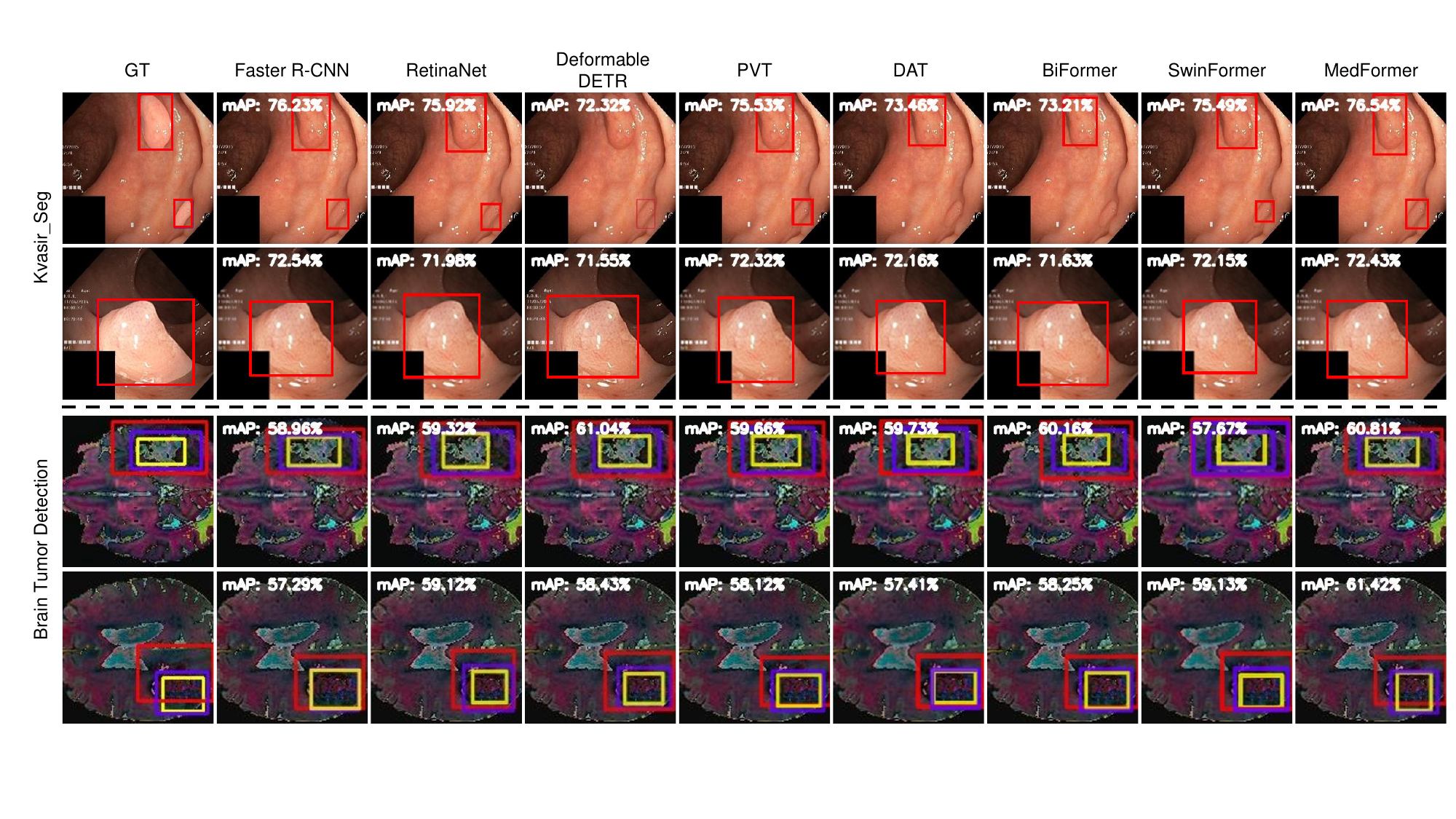}
\caption{Visualizations of object detection using various methods on the Kvasir-Seg and Brain Tumor Detection datasets. The Kvasir-Seg dataset consists of only one category, represented in \textcolor{red}{red}. The Brain Tumor Detection dataset contains three categories, represented respectively in \textcolor{red}{red}, \textcolor{yellow}{yellow}, and \textcolor{Violet}{Violet}.}
\label{fig:detection}
\end{figure*}

Fig. \ref{fig:detection} shows the object detection results of MedFormer, well-known detection methods, and popular backbone networks on the Kvasir-Seg and Brain Tumor Detection datasets. For the Kvasir-Seg dataset, both DAT and BiFormer fail to detect small objects in the first sample. On the Brain Tumor Detection dataset, SwinFormer detects the same object multiple times in the first sample, while Faster R-CNN misses one class in the second sample. In contrast, MedFormer achieves accurate object detection on both datasets, demonstrating its strong capability in handling complex detection tasks.

\section{Conclusion}
\label{sec:conclusion}
In this work, we propose a versatile medical backbone network, MedFormer. First, MedFormer adopts a four-stage pyramid architecture to reduce computational cost while generating hierarchical feature maps to enrich feature representation. Second, a dynamic DSSA mechanism is integrated into MedFormer. DSSA performs sparse selection at both the region and pixel levels, which alleviates the quadratic complexity of the self-attention mechanism and filters out noisy tokens to focus on important features.
We conduct extensive experiments on three medical image recognition tasks, and the results demonstrate superior performance compared to other state-of-the-art methods across almost all evaluation metrics.
In future work, we will continue to focus on the design of efficient Transformers to provide the medical imaging community with more lightweight and accurate solutions for medical image recognition.

\bibliographystyle{IEEEtran}
\bibliography{refer}
\end{document}